\documentclass[runningheads]{llncs}
\usepackage{graphicx}
\usepackage{siunitx}
\usepackage{color}
\usepackage{amsfonts}
\begin{document}

\title{Semantic filtering through deep source separation on microscopy images}

\author{Avelino Javer\inst{1,2}, Jens Rittscher\inst{1,2}}
\authorrunning{A. Javer and J. Rittscher}
\institute{Institute of Biomedical Engineering, Department of Engineering Science, University of Oxford, UK\\  \and 
Big Data Institute, University of Oxford, UK
\email{avelino.javer@eng.ox.ac.uk}}

\maketitle              
\begin{abstract}

By their very nature microscopy images of cells and tissues consist of a limited number of object types or components. In contrast to most natural scenes, the composition is known {\it a priori}. Decomposing biological images into semantically meaningful objects and layers is the aim of this paper. Building on recent approaches to image de-noising we present a framework that achieves state-of-the-art segmentation results requiring little or no manual annotations. Here, synthetic images generated by adding cell crops are sufficient to train the model. Extensive experiments on cellular images, a histology data set, and small animal videos demonstrate that our approach generalizes to a broad range of experimental settings. As the proposed methodology does not require densely labelled training images and is capable of resolving the partially overlapping objects it holds the promise of being of use in a number of different applications.

\end{abstract}

\section{Introduction}

In contrast to natural images, the scene composition in microscopy images is typically known. The image acquisition is done in a controlled environment where the objects of interest are highlighted by physical methods such as optical contrast, staining or fluorescence labelling. At the same time, noise, low contrast and diverse imaging artifacts can make image analysis challenging. It is not uncommon to deal with overcrowded scenes or objects with complex morphology. Additionally, many typical assumptions of natural scenes do not apply to histological images, {\it i.e.} objects can be rotated in any direction, they can be out of plane or have different appearance depending on the sample preparation.

Deep learning has been successfully adopted for microscopy image analysis, in particular for segmentation and localization tasks\,\cite{ronneberger2015u,suleymanova2018deep}, however its adoption is limited by the need for large manually annotated datasets. The annotation of microscopy images is demanding due to the large number of individuals and the heterogeneity of their shapes. Additionally, the object identification in microscopy data typically requires a level of expertise that cannot be easily outsourced. 
Deep learning has enabled a number of new approaches to image denoising and  restoration, a type of inverse problem that aims to recover an image $x$ from its degraded version $f(x)$. The general training procedure requires the collection of hundredths of pairs of noisy and cleaned images. The acquisition of clean targets  typically requires altering the imaging conditions during acquisition\,\cite{weigert2018content} making it quite demanding or in some situations nonfeasible. This limitation was partially solved, with Noise2Noise\,\cite{pmlr-v80-lehtinen18a}, where Lehtinen {\it et al.} demonstrated that it is possible to train a denoising network without the need of clean targets if noisy pairs of the same scene are given. The basic idea is that if we train a network using pairs of noisy images such as $(s + n, s + n')$ where $s$ is the same signal, and $n$ and $n'$ are the independently drawn noise, the network converges to a representation that cancels out the noise component. Nonetheless, the image restoration methods do not include any semantic information, {\it i.e.} they are not capable of separating the signal coming from the objects of interest from the signal of other objects.

In this paper, we expand on the idea of Noise2Noise to filter semantically meaningful information on microscopy images. We show that we can train a network to remove objects on training pairs where the objects change position and the scene remains static. This approach can be used to learn to identify moving objects from videos without any manual labelling. Additionally, we demonstrate that if we train a network on pairs of images where the signal from the objects of interest remains the same while the rest of the scene changes, the network learns to only retain the objects of interest. The most challenging part of this second approach is to get the required training data. For this, we use the fact that the composition of certain microscopy images is known {\it a priori} to synthesize the training set using a very simple method. We show that even if the synthetic images are far from realistic, a network trained on them is capable of producing high quality outputs. Contrary to the pixel-labelled images typically used in semantic segmentation, our semantic filtering has the advantage of retaining textural information and of resolving overlapping objects from different classes. Additionally its training does not need dense image annotations, but rather requires representative exemplars that can be extracted with traditional image segmentation methods.

In Section \ref{sec:method}, we present how we expand the Noise2Noise idea to semantically filter microscopy images and describe our synthetic image generation model. In Section \ref{sec:applications} we apply our methodology to different experimental settings. In Section \ref{sec:worm_motion} we use videos from {\it C. elegans} wild isolates to remove the animals from the background by training directly raw video data. In Section \ref{sec:microglia} we use data from fluorescent microglia cells to demonstrate that our approach is capable of removing out of focus objects while retaining complex cell morphologies. We then apply our method on two public datasets taken from the Broad Bioimage Benchmark Collection\,\cite{ljosa2012annotated}: BBBC026\,\cite{logan2016quantifying} (Section \ref{sec:hepatocytes}) where we demostrate that our method can resolve overlapped shapes, and BBBC042\,\cite{suleymanova2018deep} (Section \ref{sec:astrocytes}) where we show the same ideas can be applied to histological images. We provide a quantitative evaluation of the results.

\section{Method}
\label{sec:method}

\begin{figure}[t]
\centering
\includegraphics[width=\textwidth]{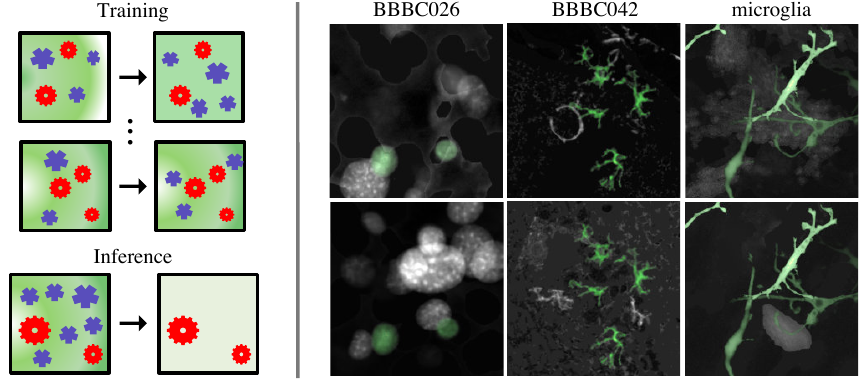}

\caption{{\bf Illustration of the training procedure}. Left, the method consists in try to make a network learn the mapping between pairs of images where the only thing in common is the target objects. Once trained, the network is then capable of outputing only the signal from the objects of interest and smoothing out the rest. Right, examples of synthetic training pairs for the different datasets. For illustration purposes the fixed parts between pair of images that correspond to the target objects are coloured in green.
}
\label{fig:explanation}
\end{figure}

We are interested in extracting data from biological samples which contain one or more entities and are set up to study phenotypical differences within a specific class or interactions between them. The object categories are referred to as $\mathcal{O}_i$. For the purpose of this paper, two objects categories are considered among all $\mathcal{O}_i$, the target category $\mathcal{O}_t$ and the rest $\mathcal{O}_u$. We will denote the background signal with $b$ and the image layer formed only by sparsely located objects from $\mathcal{O}_i$ with $\ell^i$.  Therefore an image can be modeled as $x = f(\ell^t, \ell^u, b)$. The task of interest is, given $x$, to extract the layer $\ell^t$ that contains only the pixel intensities coming from the objects of interest. 

\subsection{Training Model}

The major insight in Lehtinen {\it et al.}\,\cite{pmlr-v80-lehtinen18a} is that when a neural network is used to map one images to another, the learned representation is an average of all the plausible explanations for a given training data. In some cases, this can lead to unwanted artifacts like blurry edges. Nonetheless, if the network is trained on pairs of images with different noise realizations (Noise2Noise), {\it e.g.} same scene with different Gaussian noise, the learned representation is capable of cancel out the noise just as well as if it had been trained using cleaned targets. 

Expanding on this idea, we can use pairs of images where the layer $\ell^t$ remains the same, while we change a combination of  the layers $\ell^u$ and $b$. If the the objects in $\ell^u$ are randomly located, the network learns to smooth the background objects out while enhancing the target objects. Conversely, we can learn to remove $\ell^t$ if we train on pairs where $\ell^u$ and $b$ remains static while $\ell^t$ changes between images. As we will see in section \ref{sec:worm_motion} this later case occurs naturally on videos with moving targets and a fixed camera. For the rest of the experiments we rely on synthetic image pairs generated as explained below. For simplicity, in all our experiments we used a U-Net architecture with the same architecture as the one used by Lehtinen {\it et al.}\,\cite{pmlr-v80-lehtinen18a}. However, it is worth to remark that the approach is not limited to a specific network architecture.

\subsection{Synthetic Image Generation}
We use a simple model to create synthetic images where we assume that the image formation is additive $x = \ell^t + \ell^u + b$. Additionally, we assume that it is possible to obtain sample crops of isolated individuals $o^i_k \in \mathcal{O}_i$ and patches of the background $b$ where any pixel belonging to any $\mathcal{O}_i$ has been set to zero.  We can then create $\ell^t$ and $\ell^u$ by randomly placing $o^i_k$ from their respective classes, and $b$ from the patches. Partial overlaps between the objects is allowed (up to 50-90\% of the placed object) but we observed that complete overlaps can be detrimental leading to artifacts. Each crop is augmented using random rotations, flips, and resizes, as well as multiplying by a random constant and substracting a random constant. Additionally, the zero parts of the patch background are replaced by a random constant. 

The required crops can be extracted using traditional segmentation algorithms plus a manual or automatic filtering of representative examples. The main requirement is that the crops are exemplars of a given class. Examples of the generated training pairs are shown in Fig.\,\ref{fig:explanation}, and examples of the input crops are shown Fig.\,\,\ref{fig:gen_examples}.

\section{Applications}
\label{sec:applications}

\subsection{Learning morphology from moving objects.}
\label{sec:worm_motion}

\begin{figure}[t]
\centering
\includegraphics[width=\textwidth]{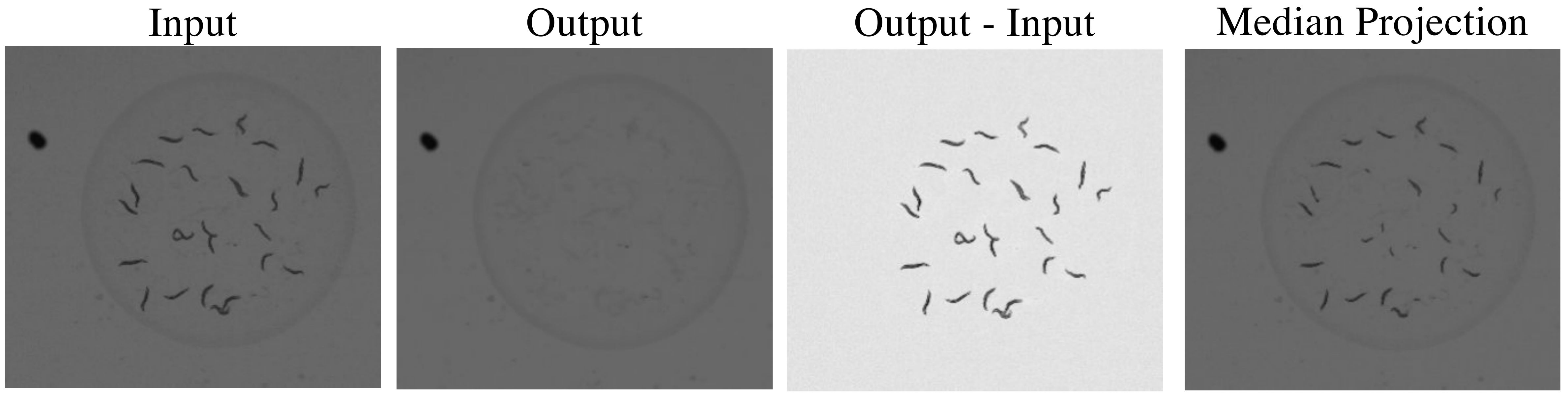}

\caption{{\bf Using consecutive frames of movies of moving objects makes possible to train a model to remove the objects in static images.} The output can be subtracted from input to recover the target objects. By comparison, the median projection along the video will fail to remove all the worms in cases where the animals motility is impaired.
}
\label{fig:embeddings}
\end{figure}

Videos taken using a static camera where only the objects of interest are moving are a natural realization of the data required to train our semantic filtering model. In this case, we can use pairs of frames separated by a fixed time lag to train a network capable of removing the target objects on single frames. This occurs because the equivalent of minimizing the L1 loss on pairs of samples with random noise is the median\,\cite{pmlr-v80-lehtinen18a}. The network effectively learns to calculate the equivalent of a median projection over a video. Therefore, as long as the median projection over a video at a given time lag results in a cleaned background, we can used those frames in our training set. The trained network output can be then subtracted from the original image in order to recover the pixels corresponding to the target objects.

We tested this approach on the set of videos of {\it C. elegans} wild isolates\,\cite{javer2018open}. We used 279 videos for training and 32 for testing. From each video we extracted five 2048x2048 frames spaced every three minutes. We trained the model on patches of 256x256 pixels on pairs of consecutive frames.  To validate the results we used as ground truth the original segmentation\,\cite{javer2018open}. The localization scores are precision (P) 0.995, recall (R) 0.999, F1-score (F1) 0.997, while the mean IoU for the whole image is 0.850. 

It is worth noting that in the training set our model is rather unnecessary, it will be much easier to calculate the median projection over each video. However, in cases were the motility of the worms is limited during inference, like in the case of mutations or drugs, the median projection will contain pixel that correspond to worms. We shown an example of this in Fig.\,\ref{sec:worm_motion} on a video of the strain {\it unc-51} that has a mutation that severely impairs the worms mobility. 

\subsection{Removing Out of Focus Cells on Microglia Cultures}
\label{sec:microglia}

\begin{figure}[t]
\centering
\includegraphics[width=\textwidth]{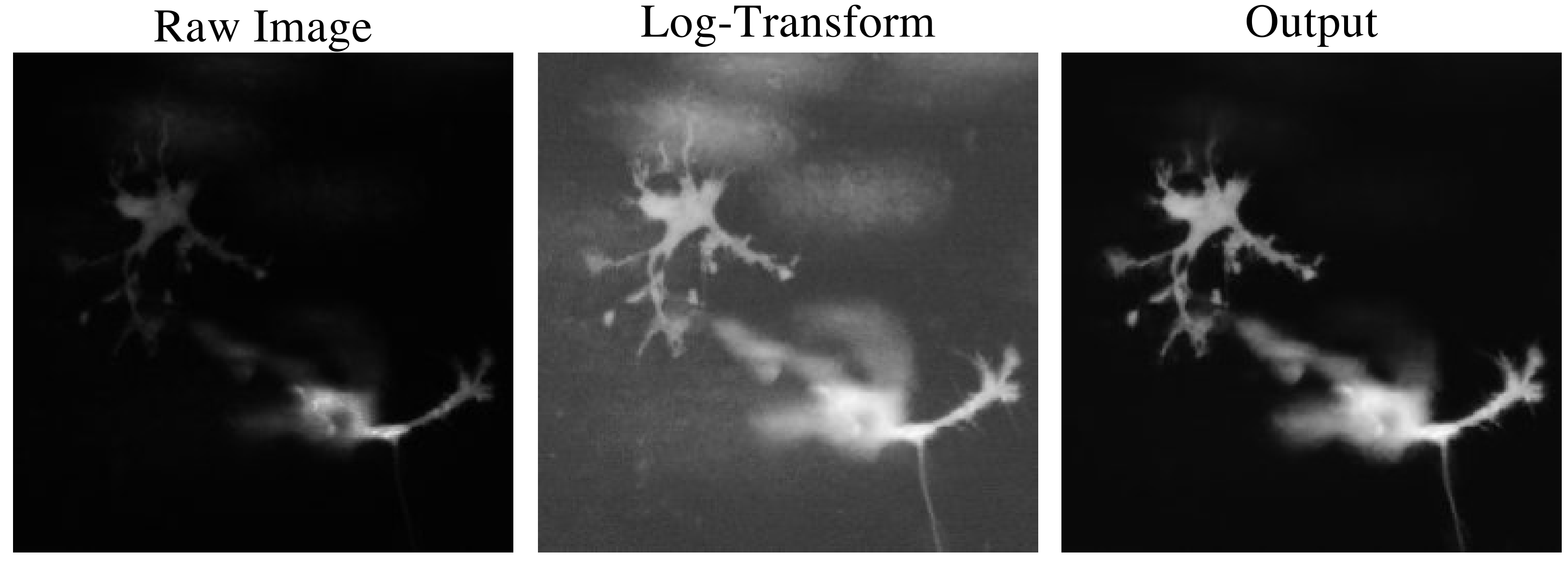}

\caption{{\bf The signal from the out of focus cells is removed while preserving the complex morphology of the microglia cells.}
}
\label{fig:microglia}
\end{figure}

The dataset consist of images of a co-culture of microglia cells and neurons. The only fluorescent cells in the images are the microglia, however, the intensity of the cell processes is much dimmer than their bodies. In order to resolve better the cell morphology we first took the log-transform of the raw images. This creates the problem of increasing the signal coming from out of focus cells. Our model is capable of cleaning the out of focus cells while retaining almost completely the cell morphology. Example of the model outputs are presented in Fig.\,\ref{fig:microglia}.

\subsection{Segmentation of co-cultures in microscopy fluorescence images.}
\label{sec:hepatocytes}

\begin{figure}[h]
\centering
\includegraphics[width=\textwidth]{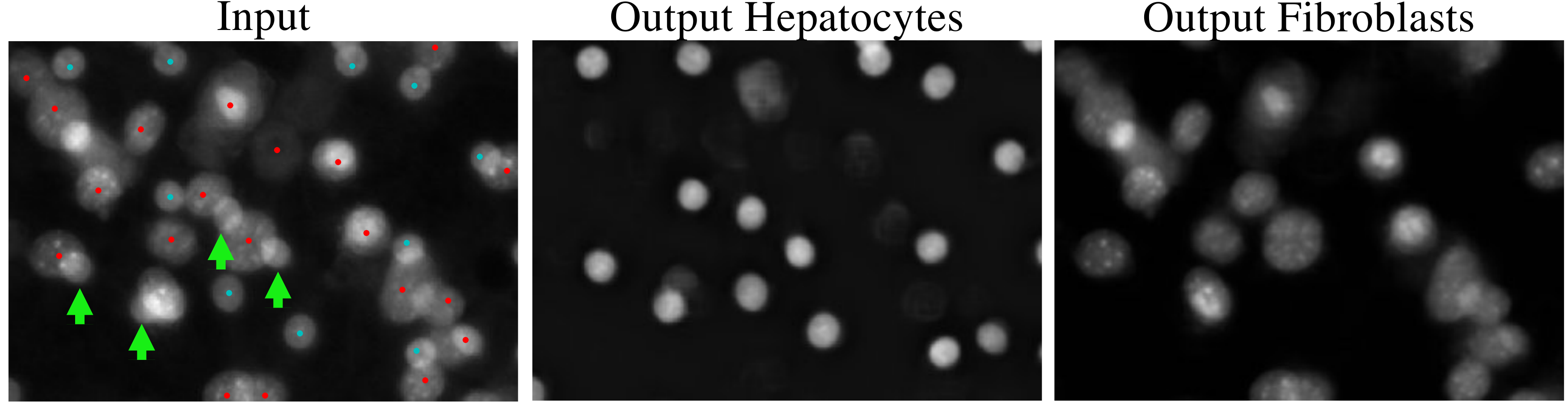}

\caption{{\bf The model is capable of resolve overlapping cells on co-cultures of hepatocyte and fibroblasts.} We display the outputs of two different models trained using either the hepatocytes or the fibroblast as the target class. The dots in the input image show the dataset annotations, fibroblast in red and hepatocytes in cyan. In this region there are four heavily overlap hepatocytes (green arrows) that were resolved by our model but are not labeled as such in the dataset. 
}
\label{fig:BBBC026}
\end{figure}

The BBBC026\,\cite{logan2016quantifying} dataset consists of images of co-cultures of hepatocytes and fibroblast taken using epifluorescence. The dataset has five hand-annotated ground truth images where each cell is labeled either as hepatocyte, fibroblast or debris. The annotations consist of a single pixel inside each object coloured according to its class. We used this information to extract the crops needed for our image synthesis model.
We train our model using 5-fold cross validation, training with four images and using the remaining one for testing. We repeat the procedure two separate times using either the hepatocytes or the fibroblasts as the target class.
The real ground truth for our method would be images with only the hepatocytes or fibroblast in separate channels. Since we do not have this data we decide to validate our results with the localization results reported in the dataset source paper\,\cite{logan2016quantifying}. Our results are for the hepatocytes (P=0.81, R=0.92, F1=0.86) compared with (P=0.94, R=0.70,  F1=0.80), and for the fibroblasts (P=0.95, R=0.96, F1=0.95) compared with (P=0.86, R=0.98,  F1=0.92). For both classes, we obtain a moderate increase in the F1-score. Interestingly, we observe that in the case of hepatocytes, the decrease of precision can be explained by apparent false positives created by overlapped cells that were not labelled in the original dataset but our network was capable of resolving (Fig.\,\ref{fig:BBBC026}).

Finally, to further corroborate our results we use the unlabelled images in the BBBC026. This images were taken using two different hepatocytes concentrations while keeping the same number of fibroblasts. Compared to the results reported in Logan {\it et al} we observe a smaller p-value among the two samples (\num{3.1e-10} vs \num{3.3e-6}), and a larger Z-factor (0.28 vs 0.16) that indicates a higher statistical power to identify between different hepatocytes concentrations.

\subsection{Segmentation on histological images.}

\begin{figure}
\centering
\includegraphics[width=\textwidth]{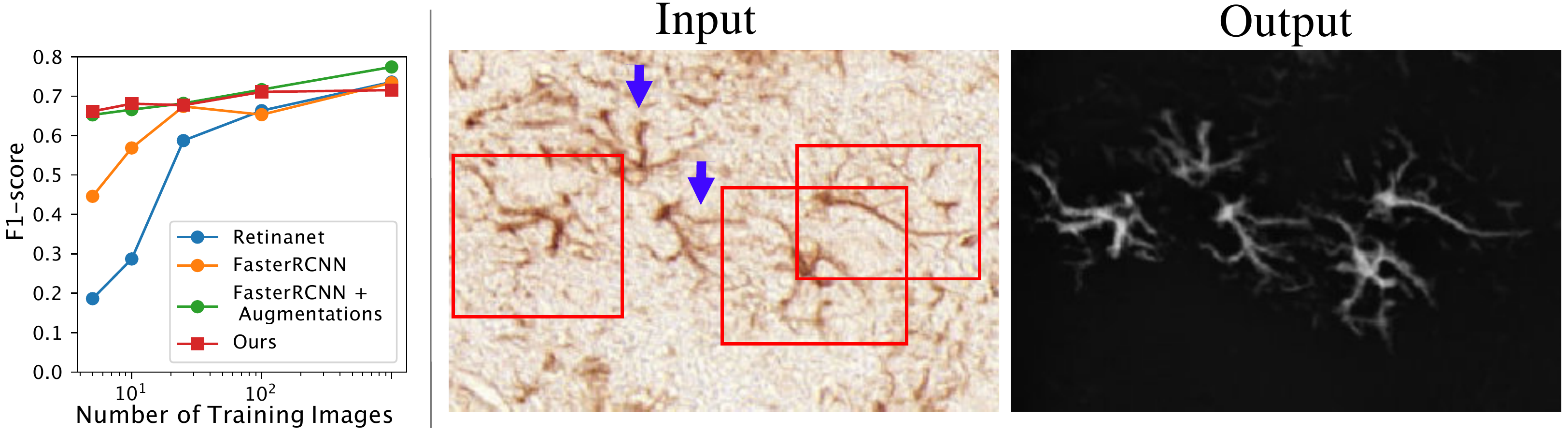}

\caption{{\bf Performance on histological images of astrocytes cells (BBBC042).} Left, our model has a performance close to the state of the art models on object detection, and shows only a small reduction of performance as the number of training images is reduced. Right, the output of our model highlights the cell morphology and could help an expert annotator to localize missing targets. In the displayed region, there are only three labeled cell (red boxes), however our model predicts two extra structures (blue arrows) that show a similar morphology to the cells.
}
\label{fig:BBBC042}
\end{figure}

\label{sec:astrocytes}
The BBBC042\,\cite{suleymanova2018deep} dataset consists of histological images from different rat brain regions stained with antibodies specific to astrocytes. The dataset is particularly challenging since the images have low contrast, diverse cell morphology and a large number of stained non-specific structures. The dataset consists of 1118 images with the location of around 15000 astrocytes marked by their bounding box.  We use 1000 images for training, 50 for validation and the rest for testing. To adapt this set into our image synthesis model, we first convert each RGB image to gray scale and calculate its complement so the background is dark and the foreground bright.

We validate the quality of our results using the localization task. The network outputs are binarized using Otsu thresholding and the bounding box calculate for each connected component removing any blob with less that 300 pixels. The assignment to the true labels is done as described by Suleymanova {\it et. al}\,\cite{suleymanova2018deep}. We compare our results with the state of the art object detection models: FasterRCNN\,\cite{girshick2015fast} and Retinanet\,\cite{lin2017focal}. We trained FasterRCNN using random crops, and horizontal and vertical flips as augmentations, however the augmentations create instability in the training for Retinanet. The results are: semantic filtering (P=0.66, R=0.78, 0.72),  FasterRCNN with augmentations (P=0.73,  R=0.82,  F1=0.77), FasterRCNN without augmentations  (P=0.80,  R=0.68,  F1=0.73) and Retinanet (P=0.81,  R=0.67,  F1=0.74). Additionally, we test how the performance  our model changes as less training data is used. The results are shown in Fig.\,\ref{fig:BBBC042} Right. When using 100 or less images, our model performance is the same as the top model, FasterRCNN with augmentations. 

The lower performance of our model when using all data could be explained by the likely presence of real cells in the background. As reported in the dataset paper\,\cite{suleymanova2018deep}, it is common that experts do not coincide in their annotations. The authors report F1 scores between 0.77-0.82 between data labelled by two different experimenters. As consequence it is likely that the dataset labels have an important number of false negatives. Since our network is not directly trained on the cell localization but rather in cleaning the background from the signal, noisy labels will affect our segmentation mask and therefore our ability to detect cells accurately. At the same time, this can be a one major advantage of our model  with respect to a network designed to only output the bounding boxes, since the network output will return a clearer outlook of the cell morphology. This output can be used to highlight to the annotator possible missing cells (Fig.\, \ref{fig:BBBC042} left).

\section{Conclusions}
Simplifying the training data acquisition is of great importance for real world settings such as high-content imaging. Our experiments demonstrate that the proposed approach can be applied to time-lapse data completely eliminating the need for any annotation. Additionally, as a result of decomposing the image into separate layers we can effectively study cells with very complex morphology without ever providing accurate dense annotations. Our work demonstrates that synthetically generated images are sufficient for training semantically meaningful mappings.  We consider that our work is of significance for a number of concrete applications in cell and tissue imaging. Going forward we will explore the application of our image synthesis paradigm in combination of different network architectures for specific tasks such as instance segmentation or pose detection.

\section{Acknowledgments}
We thank Serena Ding for providing the video of {\it C. elegans unc-51}, and Francesca Nicholls and Sally Cowley for providing the microglia data. This work was supported by the EPSRC SeeBiByte Programme EP/M013774/1. Computations used the Oxford Biomedical Research Computing (BMRC) facility. 

\bibliographystyle{splncs04}
\bibliography{egbib}

\newpage
\setcounter{page}{1}
\setcounter{figure}{0} 
\renewcommand{\thefigure}{S\arabic{figure}}%
\setcounter{section}{0} 
\renewcommand{\thesection}{S\arabic{section}}%

\section{Training Data Extraction.}
\subsubsection{Microglia.}
We generated the training data by applying a local thresholding. The connected components larger than $10^4$ pixels are considered foreground while the rest of the image is considered background. The data for $b$ is formed by the original images but with the foreground pixels set to zero. Each crop is scored by applying a Laplacian of a Gaussian filter and calculating the mean of the absolute value of the pixels inside each crop. If the score value is below a lower bound threshold, the crop is likely to contain only out of focus cells and will form part of $\mathcal{O}_u$, if it is above an upper bound threshold it should contain only cells in focus and will from part of $\mathcal{O}_t$, any value in between is discarded. 

\subsubsection{BBBC026}
To obtain the training data we first created a foreground and background mask using an Otsu thresholding on the log-transform of the original images. We then cropped the images according to the foreground connected components. A crop was included into $\mathcal{O}_t$ if it only has cells labeled as hepathocytes, in $\mathcal{O}_u$ if it only has fibroblast, or discarded if it has more than one label type. We kept 291 crops of only fibroblast and 124 crops of only hepathocytes. For $b$ we set to zero the pixels labeled as foreground in the original image.

\subsubsection{BBBC042}
In order to obtain the training data, we first converted each RGB to gray scale and calculate its complement so the background is dark and the foreground light. For $\mathcal{O}_t$, we cropped each bounding box and applied an Otsu thresholding. We only kept one connected component per crop and set the rest of the pixels to zero. For $\mathcal{O}_u$ and $b$ we first set to zero the pixels inside the labeled regions. For $\mathcal{O}_u$, we randomly selected square crops with a size between 64 to 192 pixels and apply the same procedure for $\mathcal{O}_t$. For $b$ we use local thresholding and set to zero all the foreground pixels.
\newpage
\begin{figure}

\centering
\includegraphics[width=\textwidth]{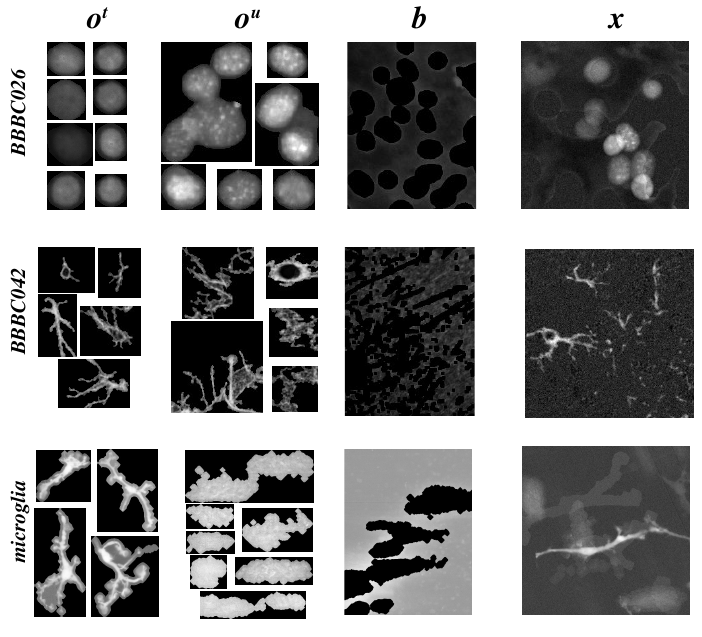}

\caption{{\bf Examples of the crops used to synthetize the images.}
}
\label{fig:gen_examples}
\end{figure}

\end{document}